\documentclass{article}

\usepackage[final]{neurips_2025}

\usepackage[utf8]{inputenc} 
\usepackage[T1]{fontenc}    
\usepackage{hyperref}       
\usepackage{url}            
\usepackage{booktabs}       
\usepackage{amsfonts}       
\usepackage{amsmath}
\usepackage{nicefrac}       
\usepackage{microtype}      
\usepackage{xcolor}         
\usepackage{graphicx}       
\usepackage{subfigure}
\usepackage[noabbrev,capitalize]{cleveref}
\usepackage{algorithm}
\usepackage[noend]{algpseudocode}
\usepackage{subcaption}

\usepackage{tcolorbox} 
\usepackage{enumitem}

\tcbset{
    colback=black!5!white, 
    colframe=black!75!black,
    boxsep=0pt, 
    left=4pt, 
    right=4pt, 
    top=4pt, 
    bottom=4pt
}
\newcommand{\fluid}{FLuID}
\newcommand{\SOLUTION}{CLIP}

\title{\SOLUTION{}: Client-Side Invariant Pruning for Mitigating \\Stragglers in Secure Federated Learning}

\author{%
  Anthony DiMaggio \\
  University of Toronto\\
  Toronto, Canada \\
  \texttt{dimaggio@cs.toronto.edu} \\
  \And
  Raghav Sharma\\
  University of Toronto\\
  Toronto, Canada \\
  \texttt{raghav@cs.toronto.edu} \\
  \And
  Gururaj Saileshwar\\
  University of Toronto\\
  Toronto, Canada \\
  \texttt{gururaj@cs.toronto.edu} \\
}

\begin{document}

\maketitle

\begin{abstract}
Secure federated learning (FL) preserves data privacy during distributed model training. However, deploying such frameworks across heterogeneous devices results in performance bottlenecks, due to straggler clients with limited computational or network capabilities, slowing training for all participating clients. This paper introduces the first straggler mitigation technique for secure aggregation with deep neural networks. 
We propose \SOLUTION{}, a client-side invariant neuron pruning technique coupled with network-aware pruning, that addresses compute and network bottlenecks due to stragglers during training with minimal accuracy loss. Our technique accelerates secure FL training by 13\% to 34\% across multiple datasets (CIFAR10, Shakespeare, FEMNIST) with an accuracy impact of between 1.3\% improvement to 2.6\% reduction.
\end{abstract}

\section{Introduction}
\label{intro}
Federated Learning (FL) enables a decentralized approach to training machine learning models on edge devices, 
allowing multiple clients to update a global model without sharing local datasets.
In each training round, clients compute gradients locally, which are aggregated by a central server and used to update a global model, which is re-distributed to participants for the next training round\cite{FedLearning}. 

\textbf{Need for Secure Aggregation.} 
While FL improves privacy by keeping client data local, model updates sent to the server can still leak sensitive information through model inversion attacks \cite{zhu2019deep,geiping2020inverting}, which show that the server can recover approximate representations of the client's training data by observing the individual client gradients.  
Secure federated learning mitigates these privacy leaks using secure aggregation (SecAgg) \cite{BonawitzSecAgg,BonawitzCCS2017SecAgg,SecAgg+}.
SecAgg ensures that the server only observes the aggregated update, not individual updates,
through secure multi-party computation (MPC) protocols that mask individual contributions, thus protecting against model inversion attacks. 
While SecAgg enhances privacy, it suffers from performance bottlenecks due to device heterogeneity in the distributed setting.


\textbf{Straggler Problem.} A central challenge in decentralized FL systems is device heterogeneity. Mobile clients vary significantly in their computational power and network bandwidth, creating ``stragglers'' or slower devices that delay training for all clients. This is because the server waits for all clients to provide their updates, before aggregating them and sending the clients the updated model. Thus, faster clients are held back by stragglers in FL, slowing down training time for all clients~\cite{reisizadeh2020stragglerresilientfederatedlearningleveraging}. 
Unfortunately, prior straggler mitigation techniques for FL are incompatible with secure aggregation.



\begin{figure*}[t!]
    \centering
    \includegraphics[scale=0.40]{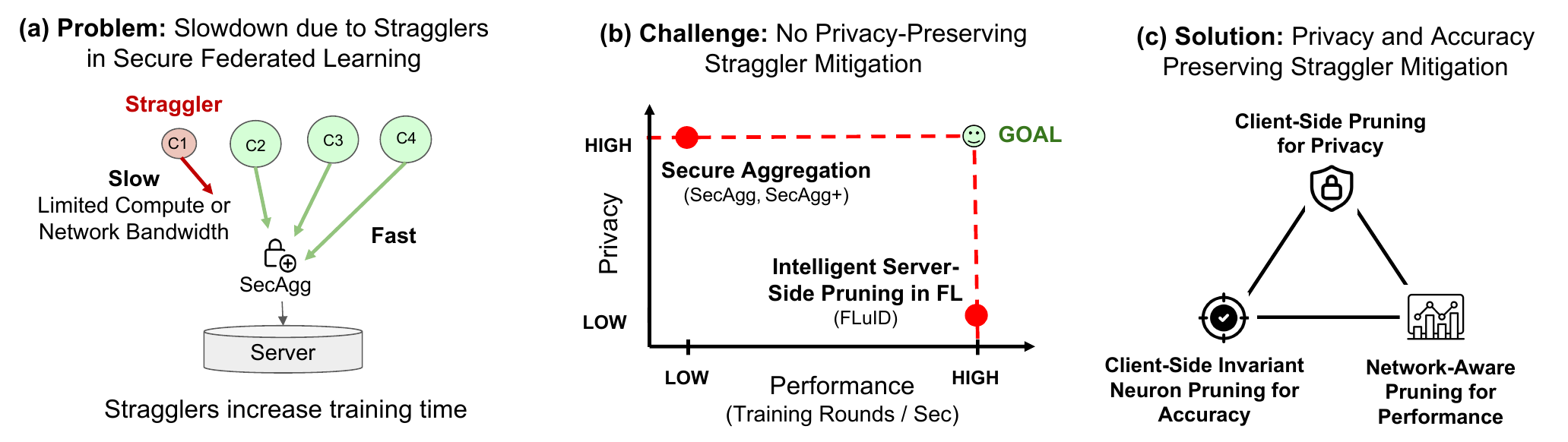}
    \caption{(a) Secure aggregation in federated learning  preserves privacy of client training data; however stragglers with slow compute or network bandwidth increase training time. (b) State-of-the-art server-side solutions for straggler mitigation are incompatible with secure aggregation and do not provide privacy. (c) Our solution enables client-side invariant neuron pruning and network-aware pruning that improves training speed while maintaining accuracy and privacy.}
    \label{fig:intro}
    \vspace{-0.1in}
\end{figure*}

\textbf{Challenge-1: Lack of Straggler Mitigation in Secure FL.} 
Prior works~\cite{wang2023fluid,reisizadeh2020stragglerresilientfederatedlearningleveraging,Distreal} propose pruning the models for straggler clients, to reduce their computational effort and equalize their training time with non-stragglers. 
Compared to leaving out the straggler's contributions,
this approach achieves higher accuracy and fairness.
State-of-the-art techniques, such as \fluid{}~\cite{wang2023fluid}, 
intelligently prune invariant neurons at the server-side to minimize accuracy impact of pruning.
Unfortunately, such approaches require the server to access individual client updates from non-stragglers to identify invariant neurons in them, which is incompatible with secure aggregation, as the server only has access to the aggregate.
Simpler client-side pruning techniques, such as Random Dropout~\cite{RandomDropout,Distreal}, 
unfortunately impact accuracy significantly. 
This leaves secure federated learning without effective straggler mitigation techniques that 
balance training time, accuracy, and privacy.

\textbf{Challenge-2: Network Stragglers in Secure FL.}
Pruning in non-secure FL~\cite{wang2023fluid, RandomDropout,Distreal} 
also mitigates network bottlenecks by reducing the size of gradient vectors transferred over the network by straggler clients. However, in secure FL,
the secure multi-party computation protocols used in SecAgg require all clients to transmit gradient vectors of identical size to ensure proper cancellation of encryption masks during aggregation. This constraint prevents the naive use of pruning techniques for reducing network overhead.
Consequently, while pruning can mitigate compute stragglers, it cannot address the bottlenecks due to network-based stragglers in secure FL.


This paper introduces the first straggler mitigation technique for secure federated learning, addressing both compute and network-related challenges. 
Our approach extends server-side pruning techniques such as invariant dropout~\cite{wang2023fluid} to a client-side solution compatible with secure aggregation (SecAgg). This reduces training time for clients, while preserving model accuracy and client privacy.

To securely address network stragglers, we identify that reducing model update sizes compromises privacy in Secure FL and hence preserve fixed-size updates by padding smaller updates from stragglers. 
Instead, we propose network-aware pruning—a technique that aggressively prunes models to reduce client training time and compensate for longer communication times. To mitigate accuracy loss from over-pruning, our method ensures stability in pruned neurons across training rounds and dynamically adjusts pruning randomness as training progresses. 
Together, this approach mitigates both compute and network bottlenecks, with accuracy impact of between 1.3\% improvement to 2.6\% reduction.



Overall, this paper makes the following contributions:

\textbf{1. Analysis of the Impact of Stragglers in Secure FL.} We provide the first systematic analysis of secure FL training times in the presence of stragglers, quantifying the effects of both compute and network stragglers on the SecAgg protocol~\cite{BonawitzCCS2017SecAgg,SecAgg+}. 

\textbf{2. Client-Side Invariant Neuron Pruning.} We propose the first client-side invariant neuron dropout technique in federated learning that is compatible with Secure Aggregation. This achieves reductions in training time for compute stragglers, reducing training time without sacrificing privacy or accuracy.

\textbf{3. Network-Aware Pruning for Stragglers.} To account for network stragglers, 
we propose more aggressive pruning to reduce client training time and compensate for the increasing network communication time. 
This reduces training time for network stragglers, 
without sacrificing privacy.

We evaluate our solution in a Secure FL setting with up to 50 clients, with the majority of non-straggler clients running at 3GHz CPU frequency with a 5G network connection,  and a minority of the clients as stragglers with a 4G connection running at 2GHz CPU frequency. 
Our technique accelerates secure FL training by 13\% to 34\% across multiple datasets (CIFAR10, Shakespeare, FEMNIST) addressing straggler overheads securely without impacting privacy, with accuracy impact of no more than 2.6\%.
\section{Related Work}

\subsection{Secure Aggregation}
Federated learning (FL) enables clients to train models locally and share gradients with a server, which aggregates them using Federated Average ~\cite{mcmahan2017FedAvg,FedLearning}. However, keeping data local is not sufficient for privacy; 
sharing plaintext gradients with the server also exposes client training data to model inversion attacks\cite{zhu2019deep, zhao2020idlg, geiping2020inverting, yin2021see, huang2021evaluating} where an attacker can recover a client's private training data from its gradients.  

Secure aggregation protocols address this using secure multi-party computation protocols (MPC) by masking individual updates with random values that cancel during aggregation, ensuring only the aggregated sum is revealed to the server \cite{BonawitzSecAgg,BonawitzCCS2017SecAgg,SecAgg+}.For each training round, the key steps in these protocols apart from the client fit include: (1) securely exchanging keys between clients, (2) generating the random mask vectors using the keys, (3) training the model and masking the gradients and transmitting them to the server, (4) the unmasking step at the server, post aggregation, where it requests the key shares from other clients for any masks that did not cancel out, allowing the server to learn the aggregated model, and (5) the server distributing the aggregated model back to clients.

\textbf{Secure Aggregation Protocols.} 
While SecAgg~\cite{BonawitzCCS2017SecAgg} first proposed secure aggregation using MPC, several variants subsequently reduce the communication overheads involved with the protocol. SecAgg+\cite{SecAgg+} replaces the fully connected communication graph among clients with a $k$-regular graph, where $k$ can be flexibly chosen to produce different performance characteristics. LightSecAgg\cite{LightSecAgg} speeds up the unmask stage by eliminating the typically used secret-share protocol and opting for mask reconstruction via encoding/decoding. FastSecAgg\cite{FastSecAgg} introduces a faster algorithm for secret-sharing, based on the fast fourier transform. 
While these methods reduce the computation or communication costs of secure aggregation, they do not address the straggler problem or device heterogeneity.
While this paper uses SecAgg+~\cite{SecAgg+}, our techniques are as applicable to other protocols.

\textbf{Straggler-aware Secure Aggregation.} CodedSecAgg encodes and strips a client's data across multiple devices to provide resiliency against stragglers   \cite{schlegel2022codedpaddedflcodedsecaggstragglermitigation}. Thus, a dropped out straggler can be compensated by contributions from non-straggler. 
However, this only addresses network stragglers, not compute stragglers, and it is only applicable to linear regression, not deep neural networks.

\subsection{Pruning for Straggler Mitigation in FL}

In a decentralized setting, clients with diverse compute or network capabilities~\cite{diaoheterofl} can slow down the training times, as the updates sent to the server are aggregated synchronously~\cite{barkai2019gap,wu2022fedadapt}. Asynchronous processing~\cite{chen2020asynchronous,chai2021fedat} or the dropping of slower clients~\cite{kairouz2021advancesopenproblemsfederated} can drop accuracy and increase convergence times.
Thus, the training round times get bottlenecked by the slowest straggler \cite{reisizadeh2020stragglerresilientfederatedlearningleveraging}. 

PruneFL~\cite{PruneFL} adjusts sub-model sizes based on varying client resources throughout training. However, it assumes uniform sub-model sizes across clients, limiting its ability to address stragglers.
%
Many recent works~\cite{Distreal,wang2023fluid,horvath2022fjord} mitigate stragglers in FL by pruning DNNs trained by the stragglers. 
Thus, slower clients contribute to the global model while training and transmitting a smaller model, 
making their round time comparable to non-stragglers and avoiding slowdown in training times.

\textbf{Non-intelligent Client-Side Pruning.} \textit{Federated Dropout or Random Dropout}~\cite{RandomDropout,Distreal} randomly choose neurons to drop out until a desired sub-model size is reached. Alternatively, \textit{Ordered Dropout}~\cite{horvath2022fjord} drops a desired number of neurons from the tail end of the filter for a given layer.
These approaches are compatible with secure aggregation as they can be implemented on the client-side.
However, the neurons thus pruned non-intelligently can degrade accuracy, making this less desirable. 


\textbf{Intelligent Server-Side Pruning.} State-of-the-art pruning techniques intelligently prune neurons to preserve accuracy. \textit{Invariant Dropout}~\cite{wang2023fluid} proposes identifying invariant neurons, whose updates from non-stragglers fall within a threshold at the server-side, and pruning these neurons from stragglers, as they have minimal impact on accuracy. 
As certain neurons train quickly and do not change much subsequently, by pruning them, Invariant Dropout minimizes the impact on accuracy. 
Unfortunately, this requires inspecting the updates of all the clients at the server-side. Thus, such intelligent straggler mitigation techniques are incompatible with secure aggregation.
Our goal is to make intelligent pruning compatible with secure FL, to speedup training times without impacting accuracy or privacy.

\section{Analysis of Straggler Impact on Training Times in Secure FL}

\begin{figure}
    \centering
    \includegraphics[width=\textwidth]{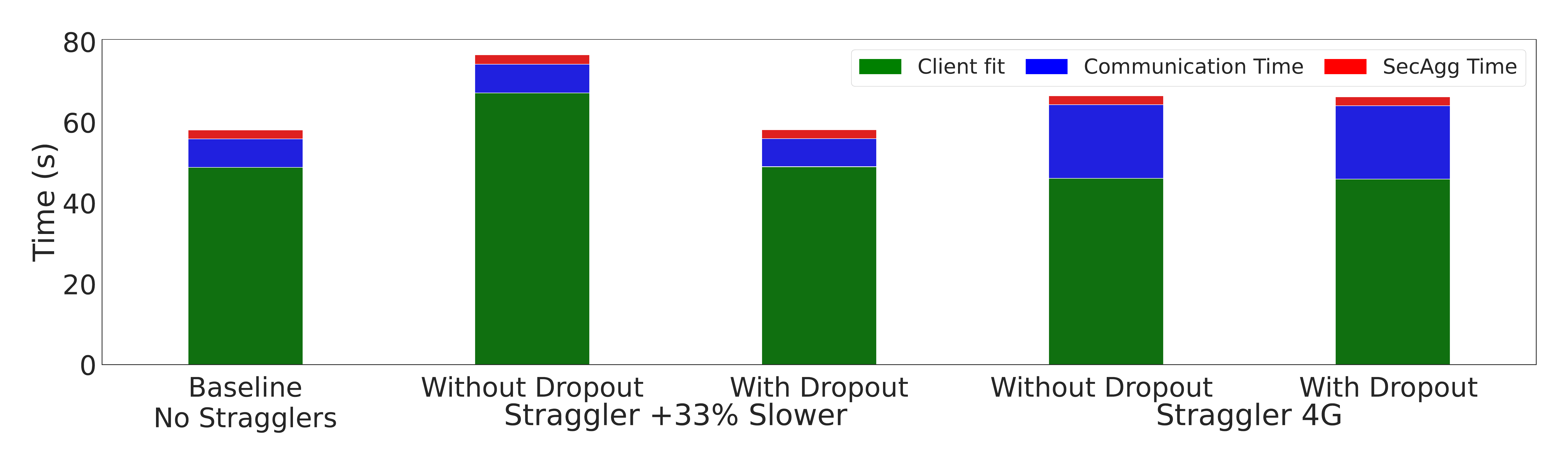}
    \caption{Time per training round for FL with secure aggregation in the presence of stragglers (clients with slower compute and lower network bandwidths).
    With compute stragglers (with 33\% slower compute), training round time increases by 24.2\%, but ideally, pruning straggler models has the potential to reduce this to within 1\%.
    For network stragglers, with less bandwidth (4G), the training time increases by 12.7\%, but dropout cannot naively address this overhead due to SecAgg constraints.}
    \label{fig:characterization-bottleneck}
\end{figure}

\label{sec:Characterization}


\textbf{Methodology.} 
We characterize 20 clients training a VGG-9 DNN model on the CIFAR-10 dataset, using the SecAgg+ implementation in the Flower FL framework~\cite{beutel2020flower}. By default, our clients have a CPU frequency of 3GHz, and a network connection equivalent to a 5G network (155 Mbps download/17 Mbps upload). We model 4 of the 20 clients as stragglers. 
We characterize two kinds of stragglers: compute stragglers -- clients with a 33\% slower compute (2GHz CPU frequency), and network stragglers -- clients with a 4G connection (27 Mbps download/7 Mbps upload).
\cref{fig:characterization-bottleneck} shows the
impact of stragglers with slower compute and network connection on training time, breaking it down into its different components: client fit (training), communication (sending and receiving gradients), and SecAgg (setup keys, mask, unmask). It demonstrates the potential to address this slowdown with an ideal dropout scheme that prunes the model trained by stragglers.

\textbf{Impact of Compute and Network Stragglers.}
Compared to a baseline without stragglers, a compute straggler (33\% slower) increases training round time by 24.2\%, due to a slower client fit time.
Similarly, the configuration with a network straggler (client with a 4G connection) causes the overall time to increase by 12.7\% compared to baseline (all 5G connections), due to an increase in the time it takes for the straggler to send and receive the gradients and update the global to and from the server.
\begin{tcolorbox}
\textbf{Observation 1:} Compute stragglers increase the overall training round time by up to 
24.2\% (due to slower client fit), while network stragglers increase it by 12.7\% (due to slower communication). 
\end{tcolorbox}

Interestingly, the secure aggregation steps (SecAgg+ setup, mask, unmask) do not incur considerable overheads, 
indicating that straggler mitigation methods in FL using pruning should be highly effective.



\textbf{Potential for Ideal Dropout.}
To evaluate the potential benefits of dropout, we implement an idealized dropout technique on the client-side, similar to prior approaches~\cite{RandomDropout,horvath2022fjord}, making the idealized assumption of no accuracy impact. 

As shown in \cref{fig:characterization-bottleneck},
dropout can reduce the overhead of compute stragglers completely, making training round time equivalent to the baseline (no stragglers), as
the straggler's client fit time no longer bottlenecks the FL rounds.
However, for network stragglers, dropout cannot provide any benefit as even after training a smaller model, the straggler has to transmit a full gradient vector, padded with 0s for the pruned layers, due to the constraints of secure aggregation.
 Thus, the slowdown due to network stragglers is difficult to mitigate with existing dropout techniques.

\begin{tcolorbox}
\textbf{Observation 2:} Client-side pruning can mitigate compute stragglers if they are made compatible with secure aggregation with minimal accuracy impact, but it cannot mitigate network stragglers.
\end{tcolorbox}

These observations motivate us to extend invariant dropout to make it compatible with secure aggregation and mitigate compute stragglers, and explore new techniques to mitigate network stragglers, as we describe next.

\section{Design of \SOLUTION: Client-Side Invariant Pruning for Stragglers in Secure FL}
\label{sec:Design}
To address the straggler problem in Secure FL without compromising accuracy or privacy, we extend prior server-side invariant pruning~\cite{wang2023fluid} fully to the client-side, ensuring compatibility with the privacy constraints of SecAgg. Our proposal, CLIP, \underline{Cl}ient-side \underline{I}nvariant \underline{P}runing, consists of the following:

\begin{enumerate}[itemsep=0pt, topsep=0pt,leftmargin=*]

\item \textbf{Straggler-Self Identification for Privacy.} We design a distributed protocol for clients to self-identify as stragglers. This decentralized approach ensures the dropout mechanism operates exclusively on the client-side, maintaining privacy by avoiding server involvement.

\item \textbf{Client-Side Invariant Neuron Dropout for Accuracy.} We propose a method to identify globally invariant neurons, to minimize the accuracy impact of pruning, with only locally available information. This reduces training time for compute stragglers while preserving model accuracy.
 
\item \textbf{Network-Aware Pruning for Performance.} To address network bottlenecks, we over-prune models on straggler clients to reduce client fit times further, compensating for increased communication times. Despite pruning, full model updates are transmitted to ensure compatibility with SecAgg, maintaining privacy while accelerating training with network stragglers.

\end{enumerate}

\begin{figure}[tb]
    \centering
    \includegraphics[width=\textwidth]{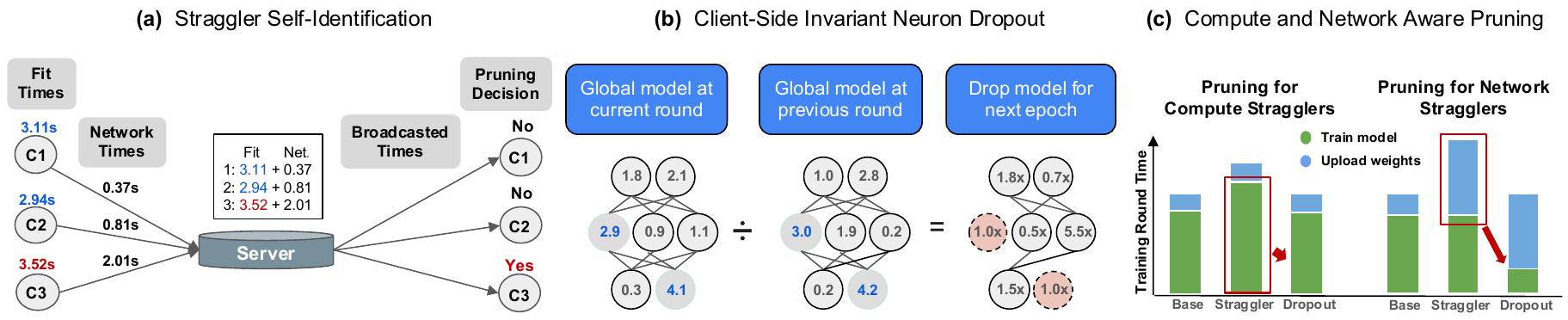}

    \caption{Overview of \SOLUTION. It consists of (a) a straggler-self identification protocol on the client-side, (b) client-side invariant neuron selection algorithm for minimal accuracy impact and maintaining privacy (compatibility with SecAgg), and (c) compute and network aware pruning for performance. 
    }
    \label{fig:clip-overview}
\vspace{-0.15in}
\end{figure}

\subsection{Client-Side Straggler Self-Identification.}
To conceal which clients prune their models and prevent the server from forcing clients to perform arbitrary amount of pruning, the clients must self-identify as a straggler. 
Fortunately, 
secure aggregation enables encrypted all-to-all communication using keys shared during setup. In \SOLUTION{}, clients reuse these keys to broadcast their fit times to others, with the server forwarding encrypted packets, as shown in \cref{fig:clip-overview}(a). By comparing received times, a client determines if it is a straggler (e.g., bottom 20th percentile) and decides on pruning.
Additionally, the server also creates an additional packet per client that contains the network times to send/receive every client's updates, and forwards these along with the compute time packets. 
These are used for network-aware pruning in \cref{sec:network_aware_pruning}.



\subsection{Client-Side Invariant Neuron Selection and Pruning}
\label{sec:client_side_design}
Intelligent pruning identifies and removes neurons that minimally contribute to model accuracy. Unlike server-side methods relying on non-straggler gradients to identify invariant neurons~\cite{wang2023fluid} to be pruned in stragglers, \SOLUTION{} operates within secure FL constraints where neither server nor client has a network-wide gradient view. Instead, it leverages temporal information, comparing global model weights from consecutive epochs to identify invariant neurons with minimal relative weight change.

\cref{fig:clip-overview}(b) shows how our technique identifies invariant neurons at the client-side. 
In each training round, the client buffers the previous epoch’s global model weights and calculates relative changes for each neuron. Neurons with changes below a threshold (e.g., 0.8$\times$ to 1.2$\times$) are deemed invariant and fit for pruning for the straggler client. This computation by the straggler is entirely server-independent.


\textbf{Invariant Neuron Selection.} First, the straggler determines the sub-model size after pruning, $p$, to achieve the desired speedup. Suppose there are $M$ stragglers, $t_k$ is the round time (computation and network times together) for the slowest non-straggler, and $T$ for the slowest non-straggler. The desired sub-model size for the $k$th slowest straggler is $p_k =  \frac{t_k}{T}$.

\begin{figure}[t] 
\noindent
\begin{minipage}[t]{0.48\textwidth}
\begin{algorithm}[H]
\scriptsize 
   \caption{Invariant Neuron Selection}
   \label{alg:invariant_selection}
    \begin{algorithmic}
       \State {\bfseries Input:} round $i$, current weights $w^{(i)}$, last weights $w^{(i -1)}$, model size $N$, sub-model size $p$
       \If{$i = 2$}
       \State Initialize $threshold = avg(\frac{w_{j}^{(i)} - w_{j}^{(i-1)}}{w_{j}^{(i-1)}})$.
       \EndIf
       \If{$i > 2$}
       \State Initialize $toDrop = []$
       \For{$j=1$ {\bfseries to} $N$}
       \If{$\frac{w_{j}^{(i)} - w_{j}^{(i-1)}}{w_{j}^{(i-1)}} \leq threshold$}
       \State $toDrop.push(j)$
       \EndIf
       \EndFor
       \If{$size(toDrop) \geq round(N * p)$}
       \State $toDrop = toDrop.sample(round(N * p))$
       \Else
       \State $alt = round(N*p) - size(toDrop)$
       \State $slack = slackNeurons(i, N, alt, prevDrop)$
       \State $toDrop.extend(slack)$
       \EndIf
       \EndIf
       \State $prevDrop = toDrop$
       \State {\bfseries Output:} $toDrop$
    \end{algorithmic}
\end{algorithm}
\end{minipage}%
\hfill
\begin{minipage}[t]{0.48\textwidth}
\begin{algorithm}[H]
   \scriptsize 
   \caption{Slack Neuron Selection}
   \label{alg:slack_selection}
    \begin{algorithmic}

       \State {\bfseries Input:} round $i$, model size $N$, slack size $S$, previously dropped neurons $prevDrop$, global accuracy history $accs$
       \State $S = round(S)$
       \If{$i > 5$}
       \State $bench = avg(accs[0, 5])$
       \State $cur = avg(accs[i -5, i])$
       \State $detPct = min(max(cur / bench, 1), 0)$
       \State $prev = prevDrop.sample(S * detPct)$
       \State $rand = seq(1..N).sample(S - size(prev))$
       \State $slack = prev.concat(rand)$
       \Else 
       \State $slack = seq(1..N).sample(S)$
       \EndIf
      \State {\bfseries Output:} $slack$
    \end{algorithmic}
\end{algorithm}
\end{minipage}
\vspace{-0.2in}
\end{figure}

For a one-layer model with $N$ neurons, such that $w_{j}^{(i)}$ is the $j^{th}$ neuron of the global model at the  $i^{th}$ round, invariance is quantified as:
\begin{equation}
    Invariance(j) = \frac{|w_{j}^{(i)} - w_{j}^{(i-1)}|}{w_{j}^{(i-1)}}
\end{equation}

After allowing two warm-up rounds, we establish an invariance $threshold$ by calculating the average invariance across all neurons and neurons meeting the invariance threshold are collected into a set $D$, $\mathcal{D} = \{ j \: | \: Invariance(j)  \leq threshold \}$.
If $|\mathcal{D}| \geq round(N * p_k)$, we have sufficient invariant neurons that can be pruned to achieve our desired sub-model size. 
If $|\mathcal{D}| < round(N * p_k)$, we need to prune additional neurons to achieve the desired sub-model size. In this case, we augment $\mathcal{D}$ by selecting $|\mathcal{D}| - round(N * p_k)$ additional ``slack'' neurons to be pruned from the remaining neurons. 

\textbf{Adaptive Slack Neuron Selection.} Slack neuron selection is critical in minimizing accuracy impact, particularly in scenarios where we need to over-prune stragglers (as described in \cref{sec:network_aware_pruning}). We select slack neurons adaptively, partially from previously dropped neurons to ensure stability, and partially at random to enable the model to learn more effectively. We observe that varying the ratio between the two based on the relative accuracy gains in successive rounds is effective as stability is preferred early on and randomness is preferred in later rounds.
Let \( accs \) represent the array of accuracy gains per round (\( accs[i] \) for round \( i \)). We compute the average of \( accs[0, 5] \) and \( accs[i-5, i] \), using their ratio to estimate how much training has progressed -- a higher ratio indicates more advanced training, prompting a higher ratio of random selection in our slack neurons. This approach stabilizes the model composition in early rounds and increases randomness in later rounds.
\cref{alg:invariant_selection} and \ref{alg:slack_selection} formalize how \SOLUTION{} selects invariant and slack neurons.

\subsection{Pruning for Network Stragglers}
\label{sec:network_aware_pruning}



\textbf{Problem.} Thus far we only discussed pruning to address increased training times caused by compute stragglers (increase in client-fit time), that were also addressed in prior works
~\cite{wang2023fluid}. 
However, our analysis in \cref{sec:Characterization} suggests that network stragglers, clients with slower network connections, also bottleneck training due to larger delays in communicating model weight updates.
Unfortunately, training a smaller, pruned model naively cannot address communication delays because full model updates must be sent to the server, padded with zeros for the pruned neurons to remain compatible with the secure aggregation protocols. 


\textbf{Our solution.} 
To mitigate the impact of increased communication times, \SOLUTION{} enables network stragglers to start weight uploads earlier by aggressively pruning their models beyond what is needed to equalize client fit times.  
This approach ensures network stragglers complete their training epochs first, freeing time to initiate communication while other clients are still training.
The resulting overlap hides longer communication times.  
\cref{fig:clip-overview}(c) illustrates how our method addresses both compute and network stragglers by pruning models even when client fit times are equal. In comparison to prior works ~\cite{wang2023fluid} that only prune to equalize client fit times, we focus on shortening the overall training epochs for network stragglers, equalizing their overall round times with non-stragglers.
The accuracy impact of over-pruning is minimized by our adaptive slack neuron selection (\cref{sec:client_side_design}), which balances the impact of random neurons (``slack'' neurons) selected during over-pruning by ensuring stability of pruned neurons across epochs, especially at the early stage of training.




\section{Privacy Arguments}
\SOLUTION{} performs straggler-aware pruning fully on the client-side, and thus is fully compatible with secure aggregation, unlike prior server-side pruning~\cite{wang2023fluid}.
Dropped layers are padded with zeros and masked using random masks compliant with the SecAgg protocol, preventing the server from inferring which layers were pruned based on network communication.
The resulting update vectors maintain identical size and entropy for both stragglers and non-stragglers.


Under an honest-but-curious threat model, the server cannot identify pruned neurons or exploit pruning for targeted model inversion attacks.
Randomized neuron selection at the client-side, which increases over training epochs, further enhances privacy.
Together, the design of \SOLUTION{} ensures that the server cannot carry out any worse model inversion attacks by specifically targeting layers of the model that have experienced dropout by the stragglers, compared to secure aggregation itself.


As secure aggregation relies on statistical privacy guarantees, where the server observes the aggregate from $N$ clients, the privacy of secure aggregation with pruning for $K$ straggler clients is lower-bounded by the privacy of secure aggregation with $N-K$ clients.
As long as stragglers are a small fraction and the set of total clients is reasonably large ($\sim$10 or more), the privacy loss is negligible.
Techniques using additive noise for differential privacy~\cite{privacySecFL} can be used with secure aggregation for stronger privacy; \SOLUTION{} is orthogonal and poses no limitations on adoption of these techniques.

\section {Evaluation}
\label{sec:Evaluation}
\subsection{Methodology}

\textbf{Model and datasets.}
Like prior FL works~\cite{wang2023fluid,edgedevicespruning}, we evaluate \SOLUTION{} with the CIFAR10 dataset using VGG-9, FEMNIST dataset using ResNet-18, and Shakespeare dataset using LSTM. These meet the resource constraints of edge devices (e.g., mobile phones) where FL may be deployed.

\textbf{Secure Federated Learning Setup.}
Our FL client and server applications are built on top of the Flower~\cite{beutel2020flower} Python library using the implementation of SecAgg+~\cite{SecAgg+} in Flower. In our experiments, we use 20 clients communicating with a single server, among which 4 clients are randomly chosen as stragglers. We divide our dataset equally into non-overlapping partitions, and each client trains on a distinct partition. 
We also perform a scalability study with 5 and 50 clients, with 20\% as stragglers.

\textbf{Evaluation Setup.}
We run our experiments on an AMD EPYC 7713 64-core system. Each client and the server process are pinned to a separate core. The client and server interact using gRPC streams established by Flower's~\cite{beutel2020flower} networking stack.
We control the network throughput for each client separately using the iproute2~\cite{iproute2} tool in Linux, which can throttle TCP/IP network traffic for the client to an arbitrary speed.
By default, non-stragglers are configured to have a CPU frequency of 3GHz (similar to a high-end Apple iPhone 15) and a network speed of 5G (155 Mbps download/17 Mbps upload). 
Stragglers are configured with a +33\% slower CPU frequency of 2GHz (similar to a medium-end Samsung Galaxy A16) and a network speed of 4G (27 Mbps download/7 Mbps upload).
The network speeds for 5G and 4G are based on the average speeds reported by OpenSignal \cite{opensignal5g,opensignal4g3g}.

\textbf{Evaluation metrics.}
 We compare \SOLUTION{} against a baseline, SecAgg implementation with no dropout, and two prior client-side dropout techniques: 1) random dropout~\cite{RandomDropout} and 2) ordered dropout~\cite{horvath2022fjord}. We compare these based on test accuracy and overall training times (wall-clock).


\subsection{Accuracy evaluations}

\begin{figure}
    \label{fig:vgg_model_size}\includegraphics[scale=0.15]{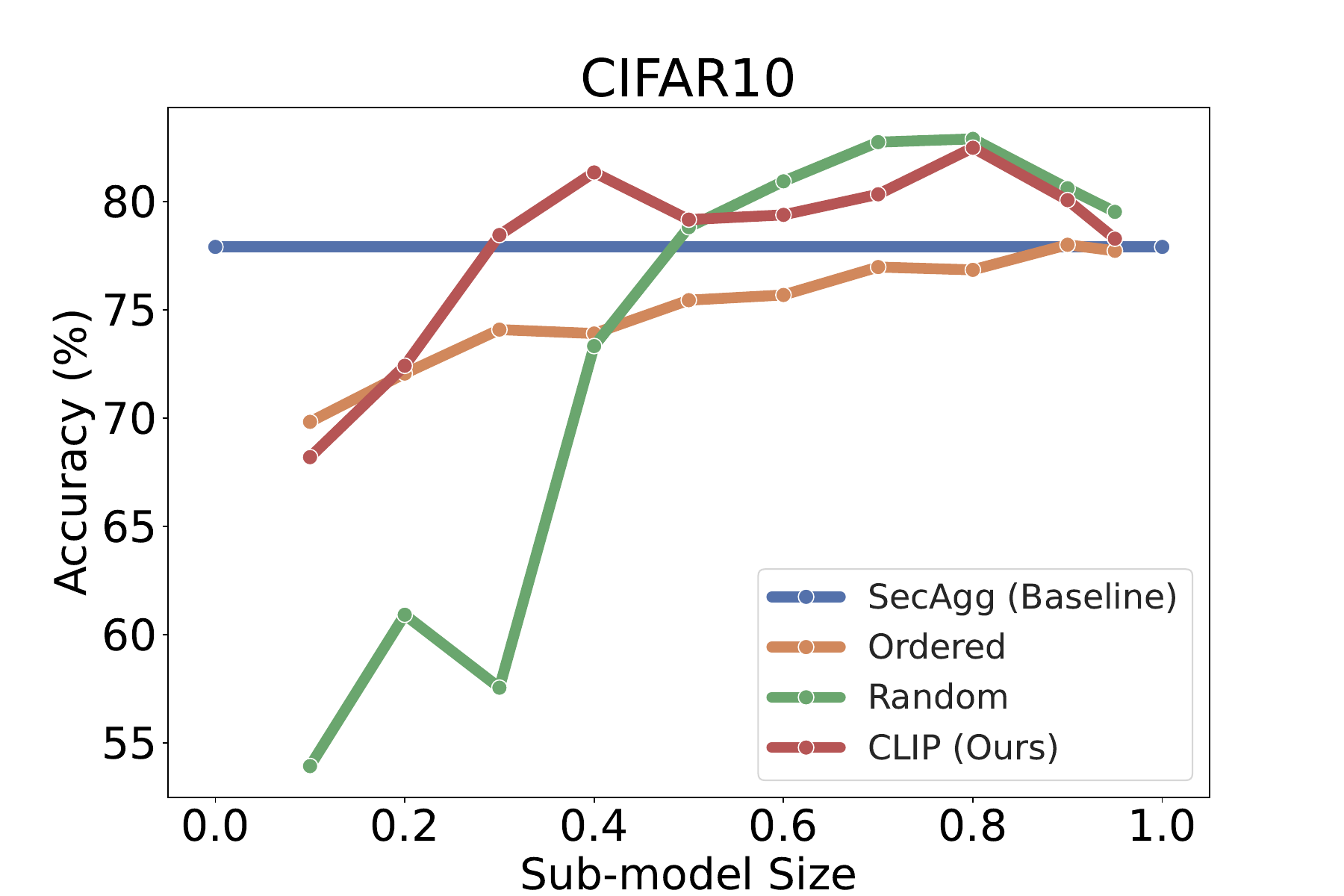}
    \includegraphics[scale=0.15]{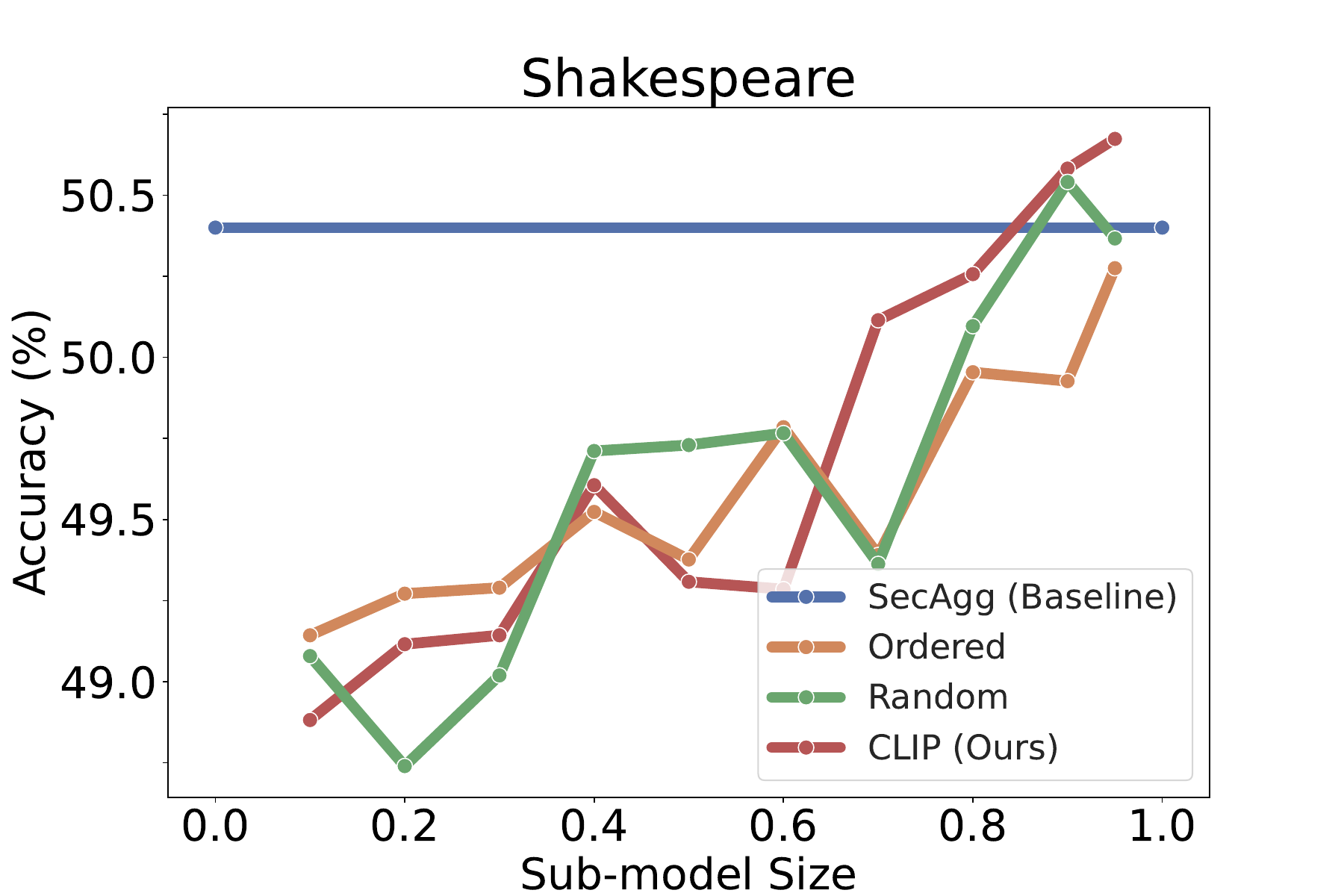}
    \includegraphics[scale=0.15]{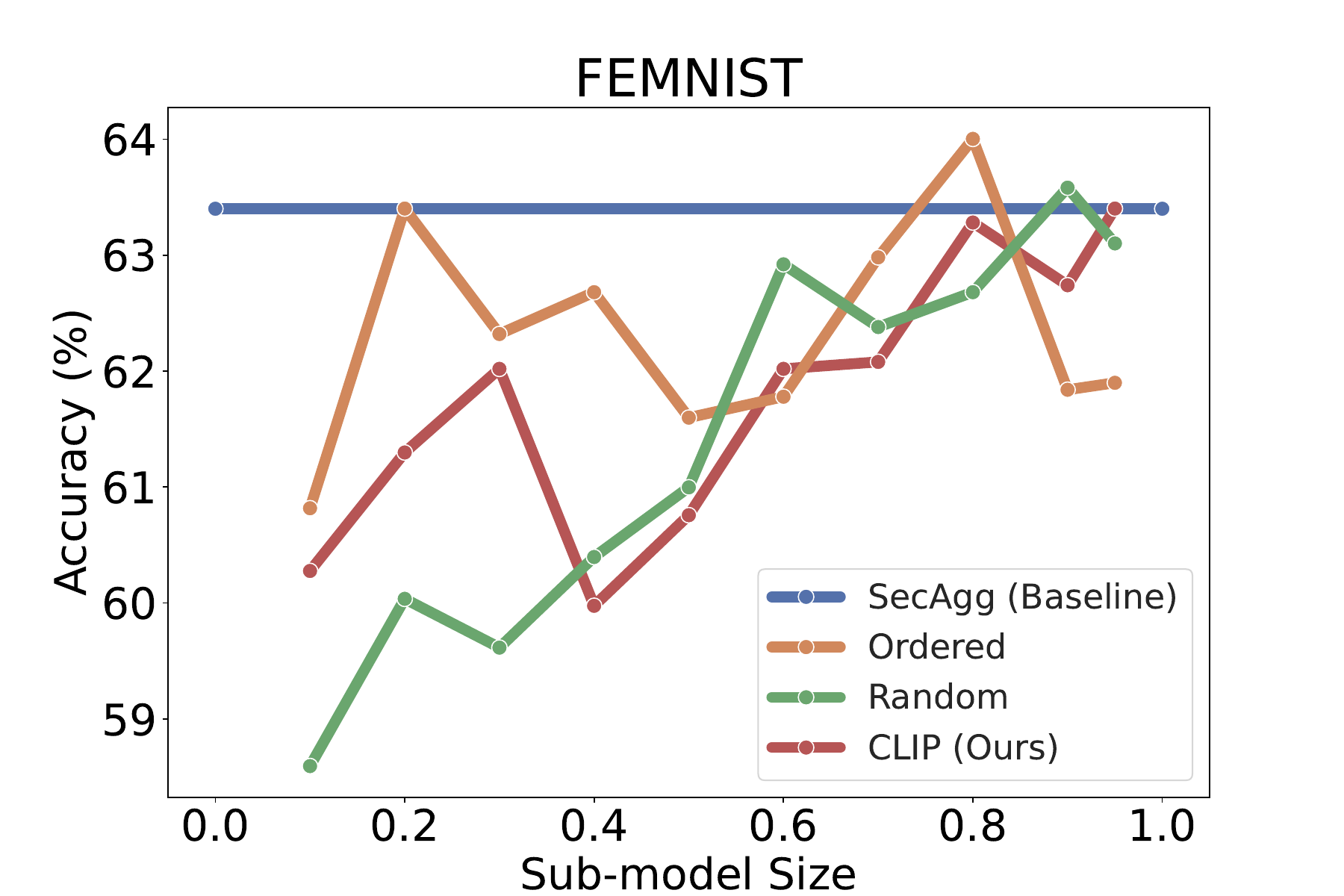}
    \caption{Accuracy versus sub-model size of \SOLUTION{} compared to Ordered and Random Dropout and SecAgg (baseline). 
    Averaged
    over 3 runs, where each run is trained for 100 rounds with 20 clients.}
    \vspace{-0.2in}
    \label{fig:vgg_accuracy_v_model_size}
\end{figure}

\textbf{Accuracy vs Sub-Model Size.}
Fig. \ref{fig:vgg_accuracy_v_model_size} compares the accuracy versus submodel size trade-off of our client-side invariant dropout versus ordered dropout, random dropout, and the baseline (SecAgg).
Across a range of sub-model sizes, \SOLUTION{} achieves as good or better accuracy compared to prior dropout techniques. 

For CIFAR10, \SOLUTION{} outperforms the baseline for submodel sizes of 0.3$\times$ to 0.9$\times$, and is better than all other dropout techniques for 0.2$\times$ to 0.5$\times$, while being comparable to Random Dropout from 0.5$\times$ to 0.9$\times$. At 0.8$\times$ sub-model size, it achieves its best model accuracy of 82.4\% accuracy, beating out baseline by 4.5\%. 
Even at sub-model size of 0.5 (the maximum pruning we permit in practice), \SOLUTION{} has an accuracy improvement of more than 1\% compared to baseline. 

For Shakespeare, \SOLUTION{} outperforms baseline and the other dropout techniques for sub-model sizes above 0.8$\times$ and is better than the other dropout techniques until the sub-model size of 0.6$\times$. For smaller sub-model sizes, it is comparable to the other dropout techniques or has slightly lower accuracy.  

For FEMNIST, \SOLUTION{} suffers an accuracy drop compared to baseline and the other dropout techniques for all sub-model sizes below 0.95$\times$. 
For the sub-model size of 0.5, which is the maximum pruning we perform in practice, \SOLUTION{} suffers an accuracy drop of almost 2\%, whereas other techniques perform slightly better -- Random is almost 1\% better at sub-model size of 0.6$\times$, while Ordered is better by almost 0.6\% at sub-model sizes of 0.7-0.8. 
However, \SOLUTION{} still provides overall speedups in training, since it trains to similar levels of accuracies much faster, as we show in \cref{sec:perf}.


\textbf{Overall Accuracy Impact.} To measure the overall accuracy impact of pruning in \SOLUTION,
we train 100 rounds with 20 clients and 4 stragglers, and record the maximum training accuracy with Baseline, \SOLUTION{} with only pruning for compute stragglers (\SOLUTION{}-C) and \SOLUTION{} for both compute and network stragglers. The difference between \SOLUTION{}-C and \SOLUTION{} shows the impact of security on accuracy, i.e., due to the constraint of secure aggregation resulting in over-pruning for network stragglers. As shown in \cref{table:accuracy} in \cref{sec:appedix-A}, \SOLUTION{} incurs an accuracy improvement of 1.3\% for CIFAR10, and a drop of 1.1\% for Shakespeare and 2.6\% for FEMNIST. CLIP-C alone has a benefit of 3.5\% for CIFAR10 and 0.4\% for FEMNIST and a drop of 0.5\% for Shakespeare. Thus, \SOLUTION{} has an accuracy impact of no more than 2.6\%, with almost 3\% attributed to security, i.e., compatibility with secure aggregation. 

\subsection{Performance Evaluations}
\label{sec:perf}

\textbf{Training Round Times.} We measure speedup in round times with \SOLUTION{} across datasets in \cref{fig:overall_eval_time}.  
We observe that depending on the straggler configuration, training round time is dominated by either the client fit and/or the network time. 
\SOLUTION{} is able to reduce the round time by 10.9\% - 31.6\% using compute-aware dropout and by an additional 2.2\% - 12.7\% using compute + network-aware dropout.

\begin{figure}
    \includegraphics[width=0.33\textwidth]{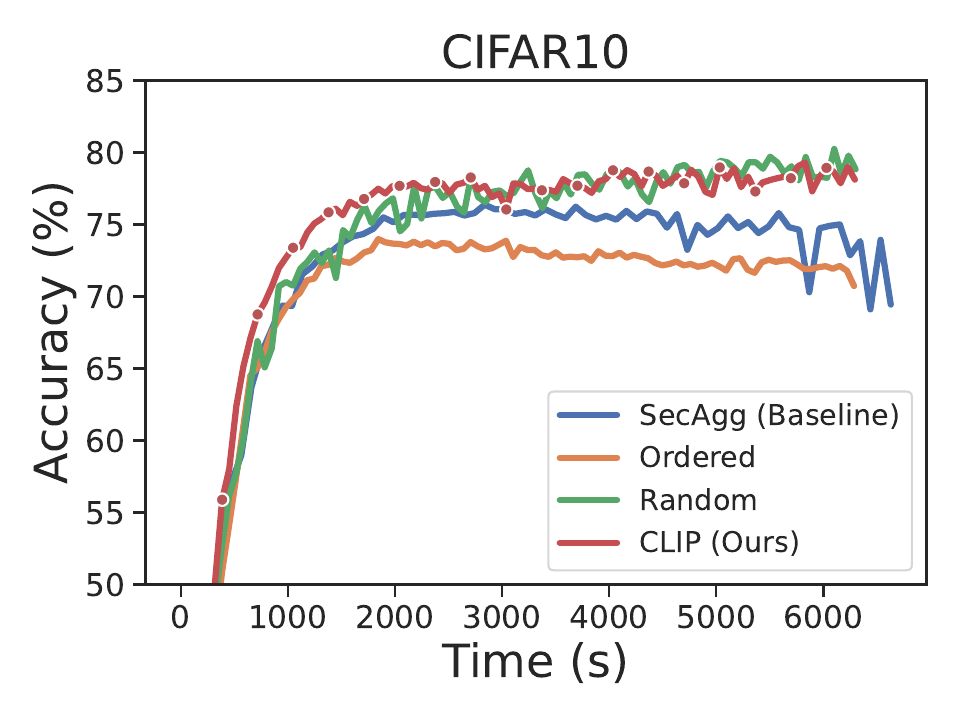}
    \includegraphics[width=0.33\textwidth]{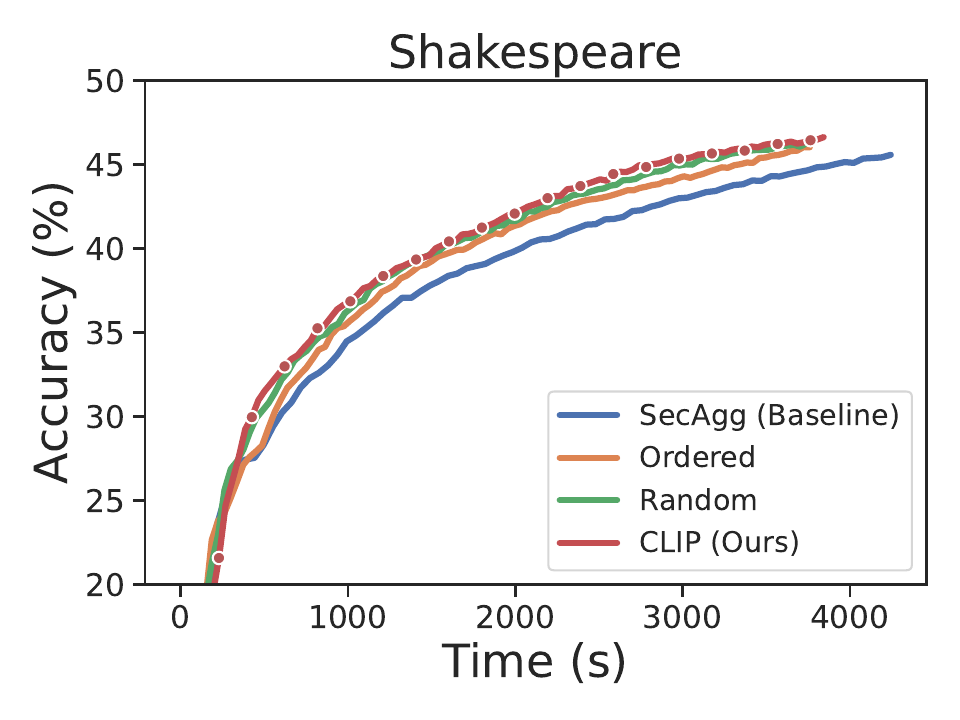}
    \includegraphics[width=0.33\textwidth]{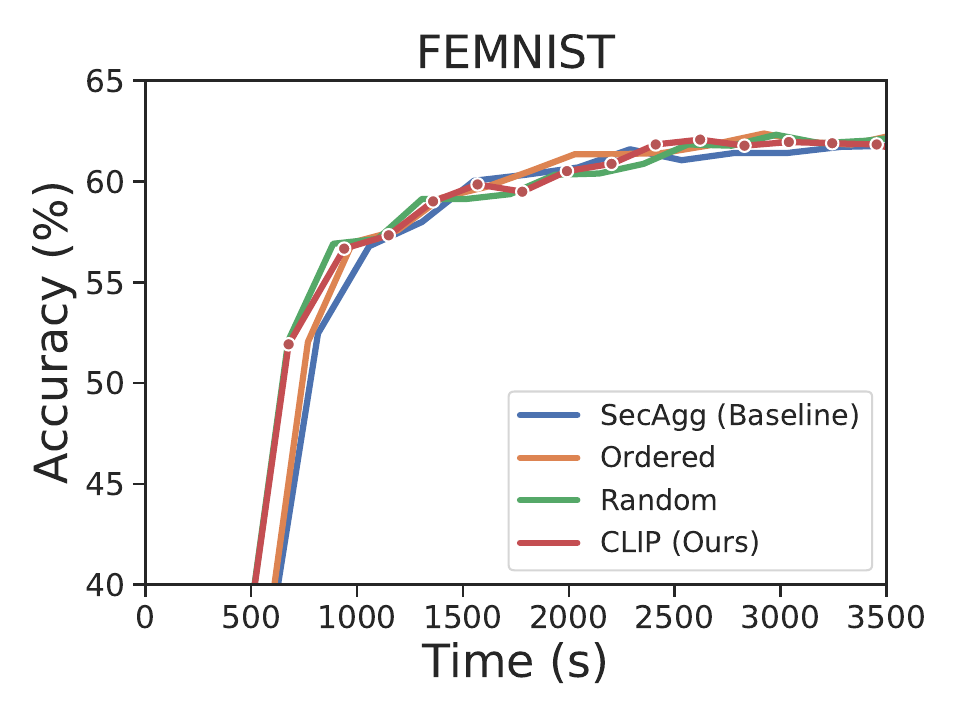}
    \caption{Accuracy versus training time for \SOLUTION{} compared to Ordered dropout, Random dropout and SecAgg (baseline) after 100 rounds, with 20 clients of which 4 are stragglers. }
    \label{fig:end_to_end}
    \vspace{-0.1in}
\end{figure}

\textbf{Accuracy vs Wall-Clock Times.}
While dropout improves training round times, it can impact accuracy per training round compared to the baseline, making it unclear whether pruning is in fact always beneficial.
To measure the effectiveness of \SOLUTION{}, we measure the accuracy versus elapsed wall clock time and measure the speedup in overall training times to reach the same level of accuracy, across all configurations. 
\cref{fig:end_to_end} shows the accuracy versus wall clock time for 20 clients across all datasets. \SOLUTION{} performs the best on CIFAR10, by training more quickly and achieving a maximum accuracy of approximately 82\%, 4\% higher than the baseline. 
Random Dropout is a bit slower than \SOLUTION, although it still achieves speedups over baseline. Ordered Dropout, however, suffers from a slowdown compared to the baseline, as its negative impact on accuracy per training round outweighs its round time improvements, leading to an overall slowdown compared to baseline. 
On Shakespeare, all dropout techniques perform significantly better than baseline, whereas on FEMNIST all dropout techniques perform similarly, narrowly outperforming the baseline. 

\textbf{End-to-End Training Time Speedup.} \cref{table:combined_speedup} provides the end-to-end speedup in training time for each technique compared to the SecAgg baseline. 
We calculate the average speedup by averaging the speedup in training time at three accuracy levels: the maximum accuracy ($m$),  $m-2.5\%$, and the $m-5\%$. Across the datasets, \SOLUTION{} achieves the highest speedup for 20 clients, achieving 30\% for CIFAR10, 27.5\% for Shakespeare, and 3\% for FEMNIST. 
For CIFAR10, the speedups for \SOLUTION{} are considerably higher than Random and Ordered, which have 12.6\% speedup and 6.7\% slowdown respectively. 
In both Shakespeare and FEMNIST, Random has a slightly lower speedup of 23.5\% and 2.1\% respectively, while Ordered has the lowest speedups with 17.1\% and 0\% respectively.  
All three dropout techniques have diminished benefits with FEMNIST as the reduced model accuracy (as shown in section 6.2) counteracts the gains in round time.

\begin{table}[htb]
\centering
\footnotesize
\caption{Average Speedup as Number of Clients Vary from 5 to 50 (20\% stragglers).}
\label{table:combined_speedup}
\begin{tabular}{c|ccc|ccc|cc}
\toprule
& \multicolumn{3}{c|}{CIFAR10} & \multicolumn{3}{c|}{Shakespeare} & \multicolumn{2}{c}{FEMNIST} \\
\cmidrule(lr){2-4} \cmidrule(lr){5-7} \cmidrule(lr){8-9}
Clients & 50 & 20 & 5 & 50 & 20 & 5 & 20 & 5 \\
\midrule
\SOLUTION{} & \textbf{21.1\%} & \textbf{30.2\%} & \textbf{33.0\%} & \textbf{14.3\%} & \textbf{27.5\%} & \textbf{17.4\%} & \textbf{3.0\%} & \textbf{24.1\%} \\
Random & 12.6\% & 18.4\% & 11.7\% & 11.0\% & 23.5\% & 11.7\%  & 2.1\% & 15.4\% \\
Ordered & -6.7\% & -6.7\% & -4.1\% & 14.0\% & 17.1\% & 13.7\% & 0.0\% & 13.2\% \\
\bottomrule
\end{tabular}
\end{table}
\textbf{Scalability.}
We also repeat the end-to-end training time speedup while varying the number of clients from 5 to 50 (maintaining 20\% of the clients as stragglers). 
\SOLUTION{} continues to outperform other techniques, achieving 21.1\% to 33.0\% speedup for CIFAR10 with 50 and 5 clients respectively, whereas Random has a 12.6\% and 11.7\% speedup respectively, while Ordered has slowdowns of 6.7\% to 4.1\%. 
These trends are similar to 20 clients for CIFAR10.
In Shakespeare, \SOLUTION{} outperforms other techniques with 14\% to 17\% speedups for 50 and 5 clients, while Random and Ordered are approximately at 11\% and 14\% speedup respectively for both number of clients.
For FEMINST with 5 clients, \SOLUTION{} has 24\% speedup compared to other techniques, higher than 20 clients, since the communication bottleneck is more limited with fewer clients, allowing pruning to provide more pronounced benefits. We find that running 50 clients is infeasible for FEMNIST with Flower since the total message size per round exceeds Flower's maximum with ResNet-18 and $>$20 clients.

\section{Limitations}
\label{sec:Limitations}
CLIP's network-aware pruning addresses network stragglers using over-pruning while padding gradients for stragglers to maintain compatibility with secure aggregation. Future work could explore advanced protocols for securely aggregating non-uniform gradient sizes to improve efficiency.
Additionally, CLIP's invariant neuron identification is designed for consistent client participation across training rounds. In dynamic environments with variable connectivity, adaptive pruning strategies could enhance robustness to intermittent participation and fluctuating bandwidth.

\section{Conclusion}
In this work, we propose \SOLUTION{}, the first straggler mitigation strategy using pruning for secure federated learning.
\SOLUTION{} is a client-side invariant neuron pruning method, which preserves SecAgg's privacy, while reducing training times at negligible accuracy cost. 
With network-aware pruning, we also compensate for the slower communication of network stragglers.
In evaluations with stragglers, \SOLUTION{} speeds up secure FL by up to 34\% while incurring negligible accuracy loss of up to 2.6\%.

\bibliographystyle{plain}
\bibliography{report/paper}

\newpage
\appendix

\section{Accuracy Impact of \SOLUTION{}}\label{sec:appedix-A}
\begin{table}[htb]
\centering
\footnotesize
\caption{Accuracy for \SOLUTION{} (Pruning for Compute $+$ Network Stragglers) with 20 clients (20\% stragglers) after 100 training rounds, compared to Baseline and CLIP-C (Pruning for Compute Stragglers Only).}
\label{table:accuracy}
\vspace{0.1in}
\begin{tabular}{c|c|c|c}
\toprule
& \multicolumn{1}{c|}{CIFAR10} & \multicolumn{1}{c|}{Shakespeare} & \multicolumn{1}{c}{FEMNIST} \\
\midrule
Baseline (No Pruning) & 77.9\% & 50.4\% & 63.4\% \\
\SOLUTION{}-C (Pruning for Only Compute Stragglers) & 81.4\% & 49.9\% & 63.8\% \\
\SOLUTION{} (Pruning for Compute $+$ Network Stragglers)  & 79.2\% & 49.3\% & 60.8\% \\
\bottomrule
\end{tabular}
\end{table}

\section{Analysis of Training Round Times with \SOLUTION{}}

\begin{figure*}[h]
    \centering
    \subfigure[CIFAR10 dataset with VGG-9 model]{\label{fig:vgg_eval_time}\includegraphics[width=0.7\textwidth]{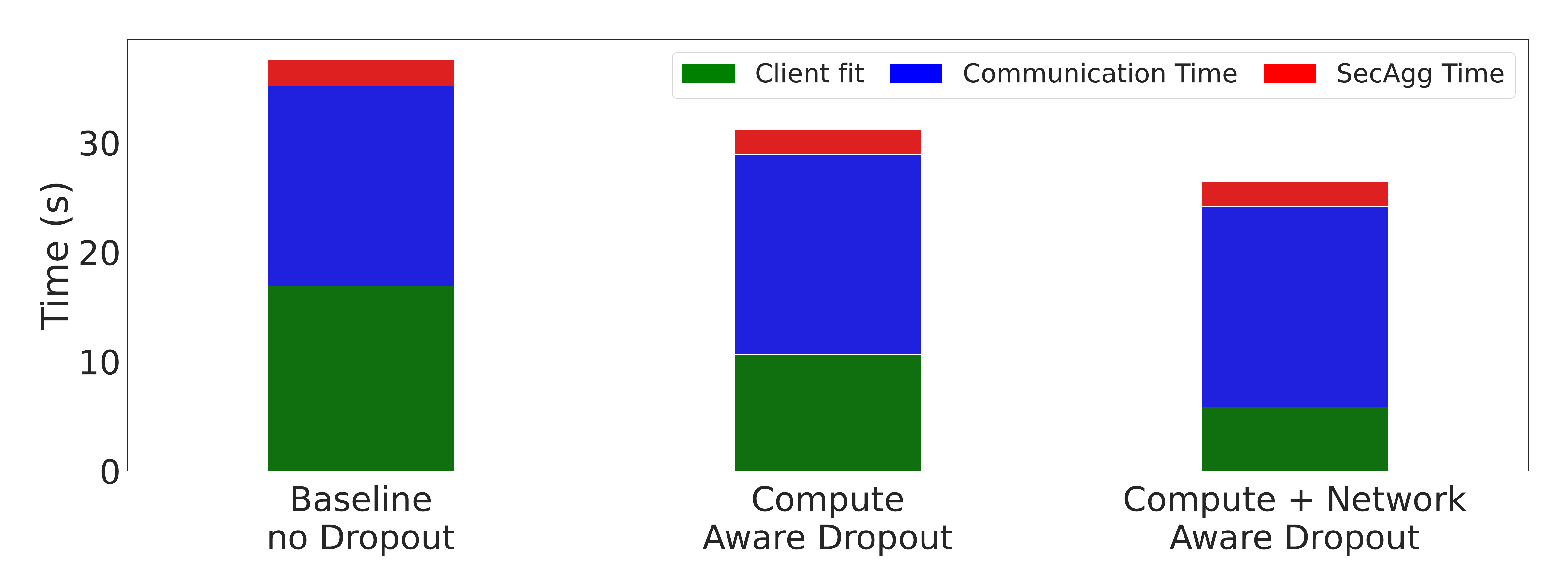}}
    \subfigure[Shakespeare dataset with LSTM model]{\label{fig:lstm_eval_time}\includegraphics[width=0.7\textwidth]{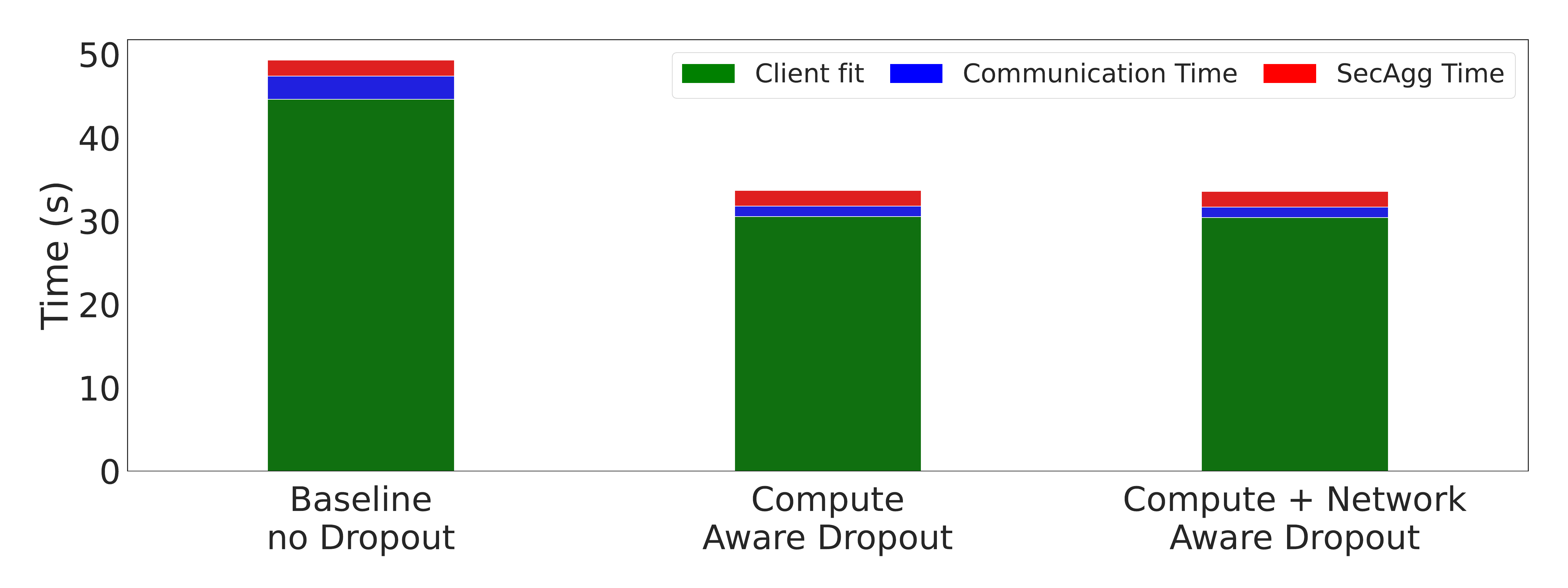}}
    \subfigure[FEMNIST dataset with ResNet-18 model]{\label{fig:resnet_eval_time}\includegraphics[width=0.7\textwidth]{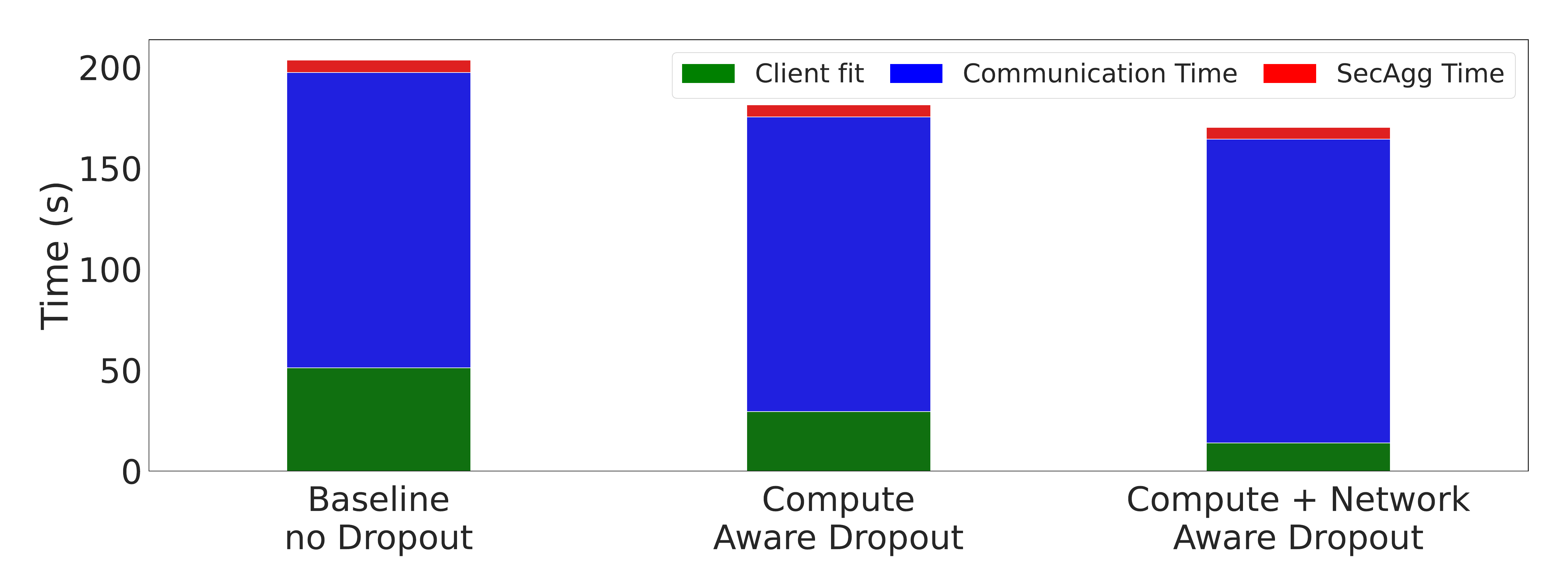}}
    \caption{Time per training round for FL with secure aggregation with and without \SOLUTION{} for configurations (a), (b), and (c), with 20 clients of which 4 are stragglers. Compute aware dropout can reduce the round time by at least 10.9\% (for FEMINIST) to at most 31.7\% (Shakespeare). With network aware dropout, the round time reduces further by at least an additional 2.2\% (Shakespeare) to at most an additional 12.7\% (CIFAR10).}
    \label{fig:overall_eval_time}
\end{figure*}

\end{document}